\documentclass{article}

\usepackage[preprint]{neurips_2026}

\usepackage[utf8]{inputenc} 
\usepackage[T1]{fontenc}    
\usepackage{hyperref}       
\usepackage{url}            
\usepackage{booktabs}       
\usepackage{amsfonts} 
\usepackage{amsmath}
\usepackage{nicefrac}       
\usepackage{microtype}      
\usepackage{xcolor}         
\usepackage{graphicx}
\usepackage{enumitem}
\usepackage{array}
\usepackage{tabularx}
\usepackage{wrapfig}
\usepackage{makecell}
\usepackage[most]{tcolorbox}
\tcbuselibrary{listings,breakable}
\newtcblisting{promptlisting}[1]{
  enhanced,
  breakable,
  colback=gray!5,
  colframe=gray!55,
  title=#1,
  fonttitle=\bfseries,
  coltitle=black,
  boxrule=0.6pt,
  arc=2pt,
  left=6pt,
  right=6pt,
  top=6pt,
  bottom=6pt,
  listing only,
  listing options={
    basicstyle=\ttfamily\footnotesize,
    breaklines=true,
    columns=fullflexible,
    keepspaces=true,
    showstringspaces=false
  },
  before skip=8pt,
  after skip=8pt
}

\title{WMLLM: Self-Evolving Optimization Agents via Predict-Then-Act World Modeling}

\author{%
  Zhongzheng Li$^{1,2,3}$,
  Qingsong Ran$^{3}$,
  Shikun Feng$^{3}$, 
  Nian Ran$^{3}$,\\
  \textbf{
  Wenhao Li$^{3}$,
  Xiaoyuan Zhang$^{3}$, 
  Yue Wang$^{3}$,
  Xiaoguang Zhao$^{1}$ }\\
  $^{1}$Institute of Automation, Chinese Academy of Sciences \\
  $^{2}$University of Chinese Academy of Sciences \\
  $^{3}$Zhongguancun Academy \\
    \texttt{zhangxiaoyuan@bza.edu.cn}
}

\begin{document}

\maketitle

\begin{abstract}
Black-box optimization problems remain challenging because of large, weakly structured, and high-dimensional search spaces. Existing methods often suffer from poor sample efficiency because they rely on direct candidate generation or trial-and-error refinement. A natural way to improve search efficiency is to use world modeling, which can help identify promising optimization directions before costly evaluation. Large language models can predict the outcomes of these candidates with nontrivial accuracy because of their implicit knowledge. Motivated by this observation, we propose WMLLM, a self-evolving optimization-agent framework based on predict-then-act world modeling. The agent first predicts promising directions and then acts to generate candidates. Combined with agentic multi-turn refinement, population-based search, and reinforcement learning, WMLLM refines both its implicit world model and its optimization strategy during search. Experiments on black-box optimization tasks, especially multi-objective molecular optimization, show that WMLLM improves sample efficiency and final optimization performance. On the multi-objective molecular optimization benchmark, WMLLM achieves state-of-the-art results under a limited evaluation budget.
\end{abstract}

\section{Introduction}

\begin{wrapfigure}{r}{0.5\columnwidth}
  \vspace{-8pt}
  \centering
  \includegraphics[width=0.48\columnwidth]{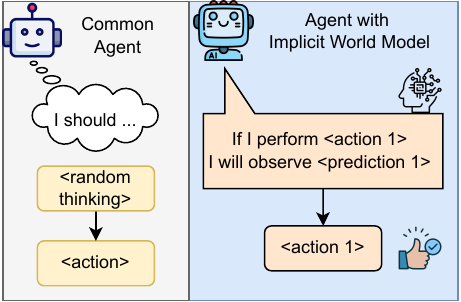}
  \caption{
Implicit world model as a predictive decision prior.
  }
  \label{f1}
  \vspace{-10pt}
\end{wrapfigure}




Recent advances in autonomous agents have sparked growing interest in applying large language models (LLMs) to structured black-box optimization problems and long-horizon decision-making problems, including scientific discovery, automated design, and embodied reasoning~\citep{zhang2025survey,weng2025deepscientist,ramrakhya2025grounding,liu2025reinforcement,zhang2025landscape}. These tasks require agents to search in complex, high-dimensional, and often discrete spaces under limited interaction budgets and sparse supervision~\citep{zhang2025agentic,xu2025incentivizing}. Representative examples include molecular design, materials discovery, program synthesis, and structured reasoning over symbolic systems~\citep{xia2025buildarena,zhang2025deepanalyze,ran2025mollm}.

Despite substantial progress, existing optimization paradigms still rely heavily on direct generation or trial-and-error exploration. Traditional reinforcement learning methods, evolutionary algorithms, and heuristic search approaches usually evaluate candidates only after execution, without explicitly anticipating their consequences~\citep{wen2025real}. This often leads to poor sample efficiency, unstable optimization behavior, and premature convergence, especially when evaluator queries are expensive and feedback is sparse~\citep{zhang2025agentrl,li2025iterative}.

A natural way to improve sample efficiency is to equip agents with world models, namely internal representations that capture how actions influence future outcomes~\citep{lecun2022path,li2025comprehensive}. Classical approaches usually instantiate world models as separate modules decoupled from the decision-making policy~\citep{schrittwieser2020mastering}. However, in high-dimensional, discrete, and partially observed optimization problems, explicit dynamics modeling is difficult to scale and may introduce extra coordination overhead~\citep{zhang2025step}. Meanwhile, LLMs have shown strong capabilities in reasoning, abstraction, and knowledge internalization~\citep{li2025deepagent}. This raises a central question: \textbf{Can a single language model improve black-box optimization by predicting the outcomes of its own actions before executing them?}
We answer this question through predict-then-act world modeling for optimization agents. Instead of learning a separate environment model, we require the LLM policy itself to explicitly predict the outcome of a candidate before acting. After evaluation, the discrepancy between predicted and observed outcomes provides a self-supervised signal for refining the agent's implicit action-outcome model. As illustrated in Figure~\ref{f1}, this changes decision making from reactive candidate generation to prediction-conditioned optimization.

To this end, we propose WMLLM, a self-evolving optimization-agent framework based on predict-then-act world modeling. WMLLM uses a single LLM to perform both outcome prediction and candidate generation. During optimization, the agent iteratively reasons over the current state, predicts the outcome of a candidate solution, receives evaluator feedback, and refines subsequent decisions based on prediction errors. This mechanism is combined with agentic multi-turn refinement and population-based search, allowing the agent to adapt its refinement depth according to external feedback while maintaining exploration over a population of candidate solutions. Through reinforcement learning, WMLLM further improves both its optimization policy and its implicit action-outcome model from interaction.
We evaluate WMLLM on a challenging multi-objective molecular optimization task~\citep{gao2022sample} and additional structured black-box optimization problems. Results show that explicit pre-action prediction improves sample efficiency and final optimization performance. In particular, WMLLM achieves state-of-the-art performance on multi-objective molecular optimization under a limited evaluation budget. Additional black-box optimization experiments and model scaling studies further show that predict-then-act world modeling provides gains complementary to larger model size and agentic refinement.

Our main contributions are summarized as follows:
\begin{enumerate}
    \item \textbf{Predict-then-act world modeling for black-box optimization.}
    We propose a self-evolving optimization-agent framework that uses pre-action prediction and prediction-observation discrepancy to improve future decisions.

    \item \textbf{Agentic tool-augmented optimization workflow.}
    We instantiate WMLLM with a multi-turn tool-use workflow that enables the agent to inspect, evaluate, and refine candidate solutions. This design enables more flexible optimization than one-shot LLM generation or conventional evolutionary operators.

    \item \textbf{Integration with population-based search and reinforcement learning.}
    We combine predict-then-act reasoning with population-based search and reinforcement learning, enabling the agent to improve its optimization policy and implicit action-outcome model through repeated interaction under limited evaluation budgets.

    \item \textbf{Empirical validation on molecular and general black-box optimization tasks.}
    We demonstrate that WMLLM improves sample efficiency and final optimization performance on multi-objective molecular optimization and additional structured black-box optimization problems.
\end{enumerate}

\section{Related Work}

\subsection{Agentic Workflows and Reflection-Based Systems}

Recent agentic workflows for large language models use structured reasoning, tool use, and self-reflection to improve performance without changing model parameters, as shown by ReAct and Reflexion-style agents~\citep{yao2022react, shinn2023reflexion}. Later systems extend this idea to long-horizon autonomy, where agents iteratively generate, execute, and refine solutions using environmental feedback, especially in scientific discovery and optimization~\citep{xia2025sr}. Related work also studies closed-loop learning through reinforcement learning or self-play, enabling agents to update internal representations from repeated interaction~\citep{chen2025internalizing}. In embodied and multimodal domains, several methods organize multi-turn reasoning around state estimation and transition prediction to improve robustness~\citep{wang2025vagen, wang2025world}.

In parallel, many works integrate LLMs with evolutionary algorithms for optimization and design. Methods such as FunSearch and EoH show that LLMs can generate heuristics or candidate solutions, while reflection and memory mechanisms improve iterative search~\citep{romera2024mathematical, liu2024evolution, ye2024reevo, liu2025mlmaster}. Recent systems further incorporate evaluator feedback into evolutionary loops, but often treat LLMs mainly as proposal modules rather than agents that refine their own predictive understanding~\citep{novikov2025alphaevolve}.

Despite these advances, most agentic workflows use reflection, tool use, and reasoning depth as procedural mechanisms, with prediction quality only indirectly linked to decisions. In contrast, WMLLM treats agentic interaction as a mechanism for world model self-evolution, where refinement is driven by prediction residuals and closed-loop optimization strengthens predictive responsibility.

\subsection{Explicit World Models and Model-Based Reinforcement Learning}

World models have long been used to improve sample efficiency and long-horizon decision-making by explicitly modeling environment dynamics. Classical model-based reinforcement learning methods learn latent transition models integrated with planning or value estimation, achieving strong results in domains with well-defined simulators such as continuous control and games~\citep{schrittwieser2020mastering, hafner2023mastering, li2025comprehensive, xing2025critiques}.

Beyond classical RL, world models have been extended to embodied AI, robotics, and web agents, where models predict task-relevant future outcomes rather than raw observations~\citep{berg2025semantic, chae2024web, feng2025evoagentselfevolvingagentcontinual}. They have also been studied in structured planning and reasoning, where dynamics are represented through symbolic abstractions or executable models for planning and verification~\citep{guan2023leveraging, lehrach2025code, pu2025one}.

These approaches usually separate world modeling from policy learning and action selection. Such modular designs can be effective, but may introduce extra modeling and coordination overhead. In contrast, our work studies whether the policy itself can act as an implicit world model by predicting decision-relevant outcomes before acting, without introducing a separate dynamics module.

\subsection{Language Models as World Models}

Recent studies suggest that large language models encode rich world knowledge and can approximate environment dynamics through next-state prediction, simulation, and counterfactual reasoning, especially in text-based settings~\citep{li2025word, wang2024can}. Benchmarks and diagnostic analyses further show that LLMs can exhibit strong short-horizon predictive ability, although their reliability and long-horizon consistency remain limited without additional structure or supervision~\citep{hu2025text2world, liu2026llm}.

Beyond passive evaluation, several approaches train or augment LLMs to predict action effects or state transitions across code, symbolic, embodied, and multimodal domains~\citep{xie2025making, sun2024enhancing, copet2025cwm, wang2025enact, chen2025planning, ge2024worldgpt}. These methods show that LLMs can function as world models when provided with suitable training signals or architectural scaffolding.

However, most existing approaches use world modeling as an auxiliary capability for evaluation, planning heuristics, or modular augmentation, rather than as a direct constraint on action selection. WMLLM instead enforces \emph{predictive responsibility}: before acting, the policy must predict the consequences of its candidate solution, and the resulting prediction error contributes directly to the learning signal. This turns implicit world knowledge into an actively refined component of decision-making for structured black box optimization.

\section{Method}

\begin{figure*}[t]
  \vskip 0.1in
  \begin{center}
  \centerline{\includegraphics[width=\textwidth]{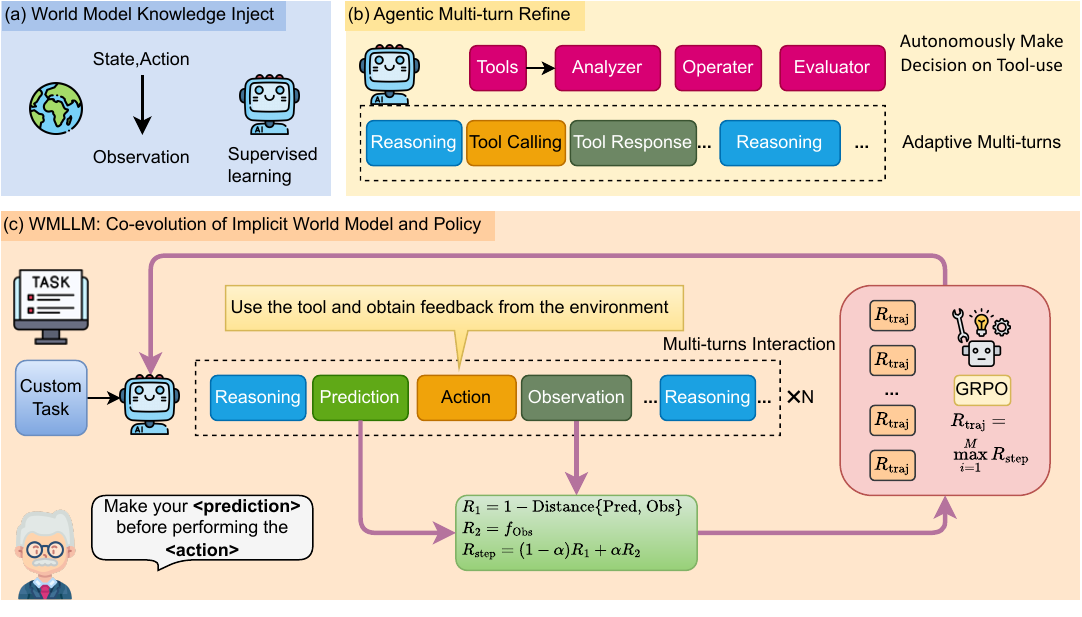}} 
    \caption{
    Overview of WMLLM.
    WMLLM co-evolves an implicit world model and policy through agentic multi-turn interaction.
    (a) World Model Knowledge Injection: the LLM is warm-started via supervised learning on state–action–observation data to encode prior environment dynamics.
    (b) Agentic Multi-turn Refinement: the LLM autonomously interleaves reasoning, tool invocation, and tool feedback with adaptive interaction depth.
    (c) WMLLM Framework: for each task, the agent iteratively predicts outcomes before acting, receives step-wise composite rewards, and is optimized using trajectory-level rewards and Group Relative Policy Optimization (GRPO).
    }
    \label{fig:overview}
  \end{center}
   \vskip -25pt
\end{figure*}

We propose \textbf{WMLLM}, a unified optimization framework that integrates population-based evolutionary search, agentic multi-turn interaction, and trajectory-level reinforcement learning within a single large language model (LLM) parameterized by $\theta$. The central idea is to treat the LLM as an adaptive agent equipped with an \textit{implicit world model}, capable of iteratively reasoning, predicting, acting, and refining candidates under sparse, black-box feedback. An overview of the complete WMLLM framework, including warm start, agentic interaction, and trajectory-level optimization, is illustrated in Figure~\ref{fig:overview}.

\subsection{Problem Formulation}

We consider black box optimization over a structured discrete search space \(\mathcal{X}\), such as molecules represented by SMILES strings or combinatorial objects represented by symbolic sequences. An external evaluator \(\mathcal{E}: \mathcal{X} \rightarrow \mathbb{R}^{K}\) maps a candidate \(x \in \mathcal{X}\) to a vector of task-specific objective values \(\mathbf{o}(x) = (o_1(x), \ldots, o_K(x))^\top\). These objectives may be competing and non-commensurable. For training and population update, we use a scalar aggregation function \(s(x) = \phi(\mathbf{o}(x))\), where \(s(x)\) denotes the normalized overall score of \(x\). The evaluator is non-differentiable and accessible only through black box queries. The goal is to discover high-quality candidates under a limited evaluation budget. Unless otherwise stated, every evaluator call made during agentic refinement is counted toward this budget.

\subsection{Population Based Evolutionary Optimization}

WMLLM adopts an evolutionary outer loop that maintains a population of candidate solutions. At generation \(t\), the population is denoted by \(\mathcal{P}_t = \{x^{(1)}, \ldots, x^{(N)}\}\), where each candidate is associated with its evaluated objective vector \(\mathbf{o}(x)\) and scalar score \(s(x)\). For each interaction with the LLM agent, we sample a parent subset \(\mathcal{P}^{\mathrm{parent}}_t \subset \mathcal{P}_t\) and provide it as part of the context. These parents ground the agent in previously explored solutions and serve as reference points for local refinement.

Conditioned on the selected parents, the LLM agent generates offspring through a multi turn trajectory. Each valid offspring is evaluated by \(\mathcal{E}\) and added to the candidate pool. To balance exploitation and diversity, the next population \(\mathcal{P}_{t+1}\) is constructed by combining high scoring candidates selected according to \(s(x)\) with diverse candidates selected by Pareto non-dominated sorting over \(\mathbf{o}(x)\). In our implementation, half of the population is selected by scalar score and half by Pareto ranking.

\subsection{Agentic Multi Turn Generation}

Given an initial context \(c_0\), including the task description, parent candidates, historical feedback, and tool specifications, WMLLM treats candidate generation as a sequential decision process. The LLM policy \(\pi_\theta(a_t \mid c_{t-1})\) produces a trajectory \(\tau = (a_1, \ldots, a_T)\), where each action \(a_t\) may contain reasoning, a candidate proposal, a prediction, or a tool invocation. The context \(c_t\) is updated after each action using the generated content and any returned tool feedback.

The agent can access task-specific tools. The analyzer provides structured information about a candidate, such as molecular fragments, substructures, or task-specific decompositions. The evaluator returns the objective vector \(\mathbf{o}(x_t)\) and the scalar score \(s(x_t)\) for a proposed candidate \(x_t\). The agent decides when to call these tools and whether to continue refinement or terminate the trajectory. This allows the trajectory length \(T\) to adapt to task difficulty and uncertainty, rather than being fixed in advance.

\subsection{Predict-Then-Act World Modeling}

WMLLM does not introduce a separate transition model or reward model. Instead, the LLM itself serves as an implicit world model by explicitly predicting decision-relevant outcomes before acting. At each refinement step, before submitting a candidate to the evaluator, the agent produces a prediction \(\hat{\mathbf{o}}_t = (\hat{o}_{t,1}, \ldots, \hat{o}_{t,K})\) for the candidate \(x_t\). After evaluation, the true outcome \(\mathbf{o}_t = \mathbf{o}(x_t)\) is observed, and the prediction discrepancy is measured by \(\delta_t = D(\hat{\mathbf{o}}_t, \mathbf{o}_t)\), where \(D(\cdot,\cdot)\) is a task-specific discrepancy function.

This predict-then-act constraint changes the role of prediction from a passive auxiliary output to an active commitment made before feedback is available. If the prediction is inaccurate, the resulting residual provides a self-supervised signal for improving later decisions. Because prediction and candidate generation are produced by the same LLM under the same context, improving prediction encourages the model to form representations that are more aligned with the evaluator. In this sense, the implicit world model is not a separate module, but a decision-relevant predictive capability embedded in the policy itself.

\paragraph{Optional warm start.}
To reduce cold start exploration, WMLLM can be initialized with supervised warm start data. Given a dataset \(\mathcal{D}\) containing candidate solutions and their evaluated outcomes, we fine-tune the model to generate both valid candidates and accurate outcome predictions in the required predict-then-act format. This stage injects task-specific prior knowledge into the agent, but it is not required by the framework. After warm start, the agent continues to improve through evaluator feedback and policy optimization.

\paragraph{Remark: Prediction and Action Coupling.}
WMLLM couples prediction and action through a shared policy. Given the interaction context \(c_{t-1}\), the LLM forms an internal state \(h_t = f_\theta(c_{t-1})\), from which it generates both the outcome prediction \(\hat{\mathbf{o}}_t\) and the candidate action \(a_t\). Since both outputs are conditioned on the same representation, the prediction discrepancy \(D(\hat{\mathbf{o}}_t, \mathbf{o}_t)\) acts as an auxiliary self-supervised signal on the representation used for action selection. When this discrepancy decreases, the model is encouraged to encode features that are more aligned with the evaluator, making subsequent candidate proposals more informed. This does not require an explicit transition model; instead, the action-conditioned predictive capability is embedded in the policy itself. This coupling provides a practical mechanism through which predict-then-act training improves optimization decisions.

\begin{figure*}[t] 
  \vskip 0.2in
  \begin{center}
\centerline{\includegraphics[width=\textwidth]{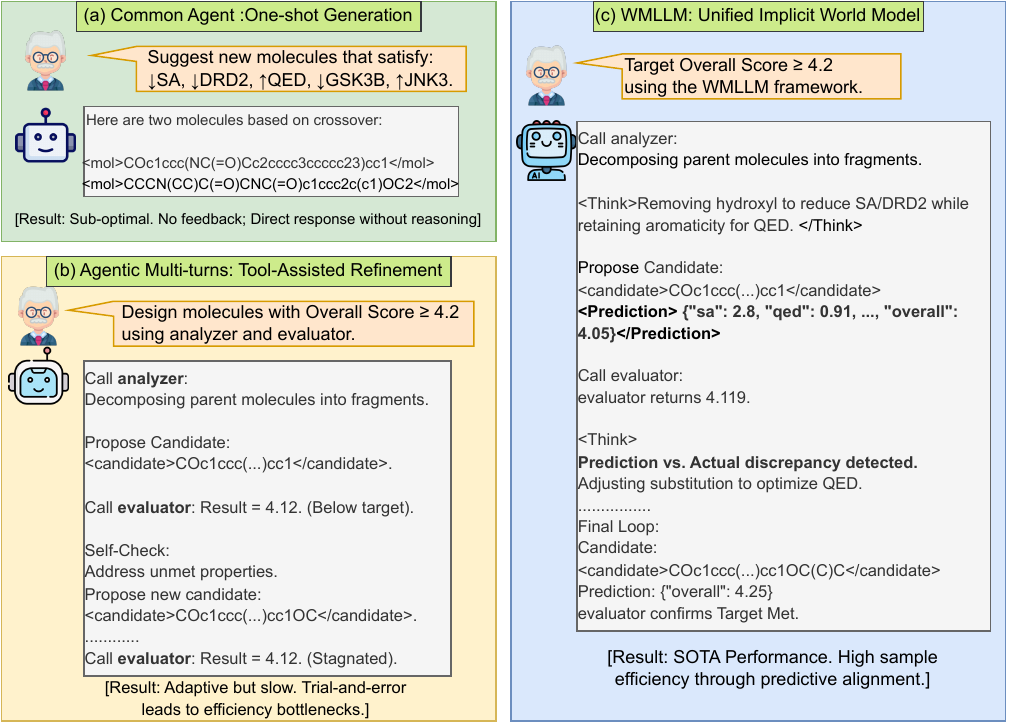}} 
    \caption{
Comparison of generation paradigms for multi-objective molecular optimization.
(a) One-shot generation directly maps prompts to candidates without feedback.
(b) Agentic multi-turn refinement iteratively improves candidates via tool-assisted trial-and-error.
(c) WMLLM integrates prediction, action, and feedback into a unified implicit world model, enabling more sample-efficient and goal-directed optimization.
    }
    \label{f3}
  \end{center}
  \vskip -20pt
\end{figure*}

\subsection{GRPO Training}

\paragraph{Trajectory-level reward.}
For each evaluated candidate \(x_t\), we compute an evaluator reward \(r_t^{\mathrm{eval}} = s(x_t)\) and a prediction reward \(r_t^{\mathrm{pred}} = 1 - D(\hat{\mathbf{o}}_t, \mathbf{o}_t)\). These two signals are combined into a step reward \(r_t = (1-\alpha) r_t^{\mathrm{eval}} + \alpha r_t^{\mathrm{pred}}\), where \(\alpha \in [0,1]\) controls the strength of the prediction signal. Rather than summing rewards over all refinement steps, we define the trajectory reward as \(R(\tau) = \max_{t \leq T} r_t\). This matches the optimization objective: a trajectory is successful if it discovers at least one high-quality candidate with reliable predicted outcomes. For reporting and population update, we separately track the best evaluator score \(\max_{t \leq T} s(x_t)\).

\paragraph{Group relative policy optimization.}
For each prompt, corresponding to a sampled parent set and task context, we sample \(K\) trajectories \(\{\tau_i\}_{i=1}^{K}\). Let \(S_i = R(\tau_i)\) denote the reward of trajectory \(\tau_i\). GRPO computes a group-normalized advantage \(A_i = (S_i - \mu)/(\sigma + \epsilon)\), where \(\mu = K^{-1}\sum_i S_i\), \(\sigma = \mathrm{Std}(\{S_i\}_{i=1}^{K})\), and \(\epsilon\) is a small constant for numerical stability. The policy is optimized with \(\mathcal{L}_{\mathrm{GRPO}} = - \mathbb{E}_i [ A_i \log \pi_\theta(\tau_i \mid c_0) ]\), where \(\log \pi_\theta(\tau_i \mid c_0)\) is the log probability of the sampled trajectory under the current policy.

This group-relative objective avoids training a separate value model and is well suited to sparse black box optimization rewards. Since advantages are normalized among trajectories sampled from the same prompt, the model is encouraged to prefer trajectories that outperform alternative refinements of the same parent set. The trajectory-level formulation also naturally supports variable length agentic interactions, because trajectories with different numbers of refinement steps can be compared through the same final reward.

\section{Experiments}

We evaluate WMLLM on multi-objective molecular optimization and additional black box optimization tasks. The experiments examine whether WMLLM improves optimization under a fixed evaluation budget, whether the gains come from predict-then-act world modeling rather than model scale or agentic refinement alone, and whether the same mechanism transfers beyond molecular design.

\subsection{Experimental Setup}

\begin{wrapfigure}{r}{0.58\textwidth}
  \vspace{-8pt}
  \centering
  \includegraphics[width=0.56\textwidth]{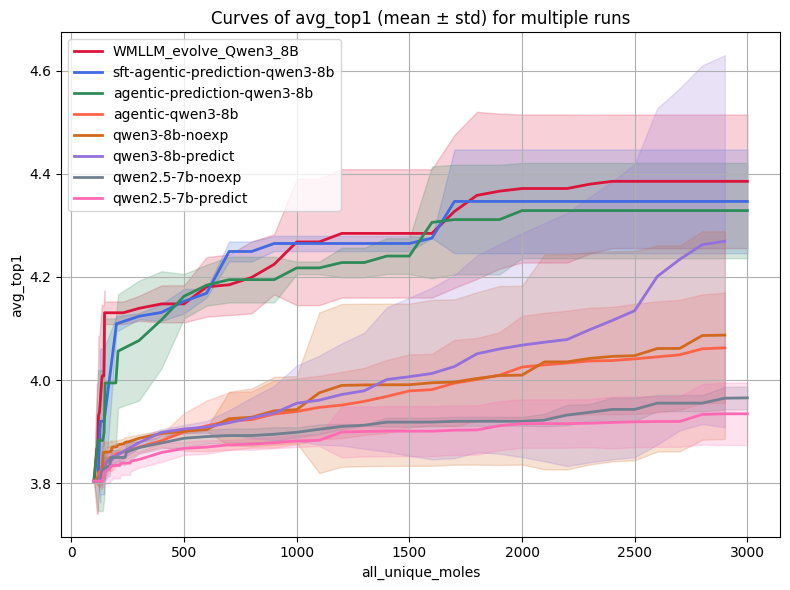}
  \caption{Optimization curves on multi-objective molecular optimization.}
  \label{f4}
  \vspace{-20pt}
\end{wrapfigure}

The main benchmark is multi-objective molecular optimization~\citep{gao2022sample,ran2025mollm}. The task is to design molecules that minimize synthetic accessibility (SA), DRD2, and GSK3B while maximizing QED and JNK3. Each molecule is represented by a SMILES string, and a black box evaluator returns a five-dimensional property vector that is aggregated into a normalized overall score. Additional details on the benchmark, evaluator, and experimental protocol are provided in Appendix~\ref{app:implementation}.


All methods in Tables~\ref{model-performance-table} and~\ref{ablation} are evaluated under the same budget of \(3{,}000\) unique molecule evaluations. Invalid molecules also consume evaluation budget. We report Avg. Top-1, Avg. Top-10, Top-10 AUC, validity, uniqueness, and diversity. Avg. Top-1 and Avg. Top-10 measure final solution quality, while Top-10 AUC measures sample efficiency across the search process. Results are reported as mean \(\pm\) standard deviation over 5 random seeds.

\begin{table}[t]
  \caption{Main comparison on multi-objective molecular optimization under \(3{,}000\) unique molecule evaluations.}
  \label{model-performance-table}
  \centering
  \small
  \setlength{\tabcolsep}{2.5pt}
  \renewcommand{\arraystretch}{1.08}
  \newcolumntype{Y}{>{\centering\arraybackslash}X}
  \begin{tabularx}{\textwidth}{@{}lYYYYYY@{}}
    \toprule
    Model & \makecell{Avg.\\Top-1} & \makecell{Avg.\\Top-10} & \makecell{Top-10\\AUC} & Validity & Uniqueness & Diversity \\
    \midrule
    WMLLM-Evolve & \textbf{4.385$\pm$0.130} & \textbf{4.255$\pm$0.084} & \textbf{4.124$\pm$0.042} & 0.642$\pm$0.013 & \underline{0.944$\pm$0.014} & 0.476$\pm$0.198 \\
    OpenEvolve & 4.123$\pm$0.067 & 4.069$\pm$0.132 & 3.661$\pm$0.081 & 0.910$\pm$0.020 & 0.504$\pm$0.099 & 0.493$\pm$0.046 \\
    MOLLEO & 4.152$\pm$0.055 & 4.058$\pm$0.023 & 3.822$\pm$0.090 & 0.940$\pm$0.008 & 0.535$\pm$0.084 & \textbf{0.657$\pm$0.014} \\
    GFlowNet & \underline{4.214$\pm$0.222} & \underline{4.175$\pm$0.193} & \underline{4.038$\pm$0.155} & \underline{0.998$\pm$0.000} & 0.345$\pm$0.005 & 0.601$\pm$0.055 \\
    GB-GA & 4.089$\pm$0.079 & 3.976$\pm$0.044 & 3.854$\pm$0.081 & \textbf{1.000$\pm$0.000} & 0.857$\pm$0.022 & 0.629$\pm$0.066 \\
    REINVENT & 4.176$\pm$0.146 & 4.096$\pm$0.243 & 3.862$\pm$0.076 & 0.979$\pm$0.002 & 0.650$\pm$0.072 & \underline{0.649$\pm$0.056} \\
    DyMol & 4.195$\pm$0.090 & 4.157$\pm$0.122 & 3.940$\pm$0.054 & \textbf{1.000$\pm$0.000} & \textbf{0.988$\pm$0.005} & 0.566$\pm$0.081 \\
    \bottomrule
  \end{tabularx}
  \vskip -0.1in
\end{table}

\subsection{Main Results on Molecular Optimization}

Table~\ref{model-performance-table} compares WMLLM with representative molecular optimization baselines under the same \(3{,}000\)-evaluation budget. WMLLM-Evolve achieves the best Avg. Top-1, Avg. Top-10, and Top-10 AUC, indicating that it not only finds stronger individual molecules but also maintains a better set of top candidates throughout optimization. This is especially important because Top-10 AUC reflects the anytime behavior of the optimizer rather than only the final result.

Compared with OpenEvolve, WMLLM-Evolve improves Avg. Top-1 from \(4.123\) to \(4.385\) and Top-10 AUC from \(3.661\) to \(4.124\). This suggests that the gain is not only due to using an LLM as an evolutionary proposal generator, but also comes from the predict-then-act interaction and policy optimization mechanism. Compared with strong molecular optimization baselines such as MOLLEO~\citep{wang2024efficient}, GFlowNet~\citep{kim2024genetic}, GB-GA~\citep{jensen2019graph}, REINVENT~\citep{olivecrona2017molecular}, and DyMol~\citep{shin2024dynamic}, WMLLM obtains higher top-score metrics while remaining competitive in uniqueness. Although its validity is lower than some graph-based or domain-specific methods, all invalid candidates are counted under the same evaluation budget, and WMLLM still achieves the best optimization quality. This indicates that WMLLM trades some validity for broader exploration while still discovering higher-scoring molecules within the fixed budget.
Figure~\ref{f4} plots Avg. Top-1 as a function of unique evaluated molecules. WMLLM rises faster than one-shot and agentic baselines, indicating better sample efficiency in the early and middle stages of search. Agentic refinement without prediction improves more slowly and saturates earlier, suggesting that multi-turn tool use alone can still behave like structured trial and error when the agent is not required to anticipate outcomes before acting.

\subsection{Ablation Study}

\begin{wrapfigure}{r}{0.58\textwidth}
  \vspace{-8pt}
  \centering
  \includegraphics[width=0.56\textwidth]{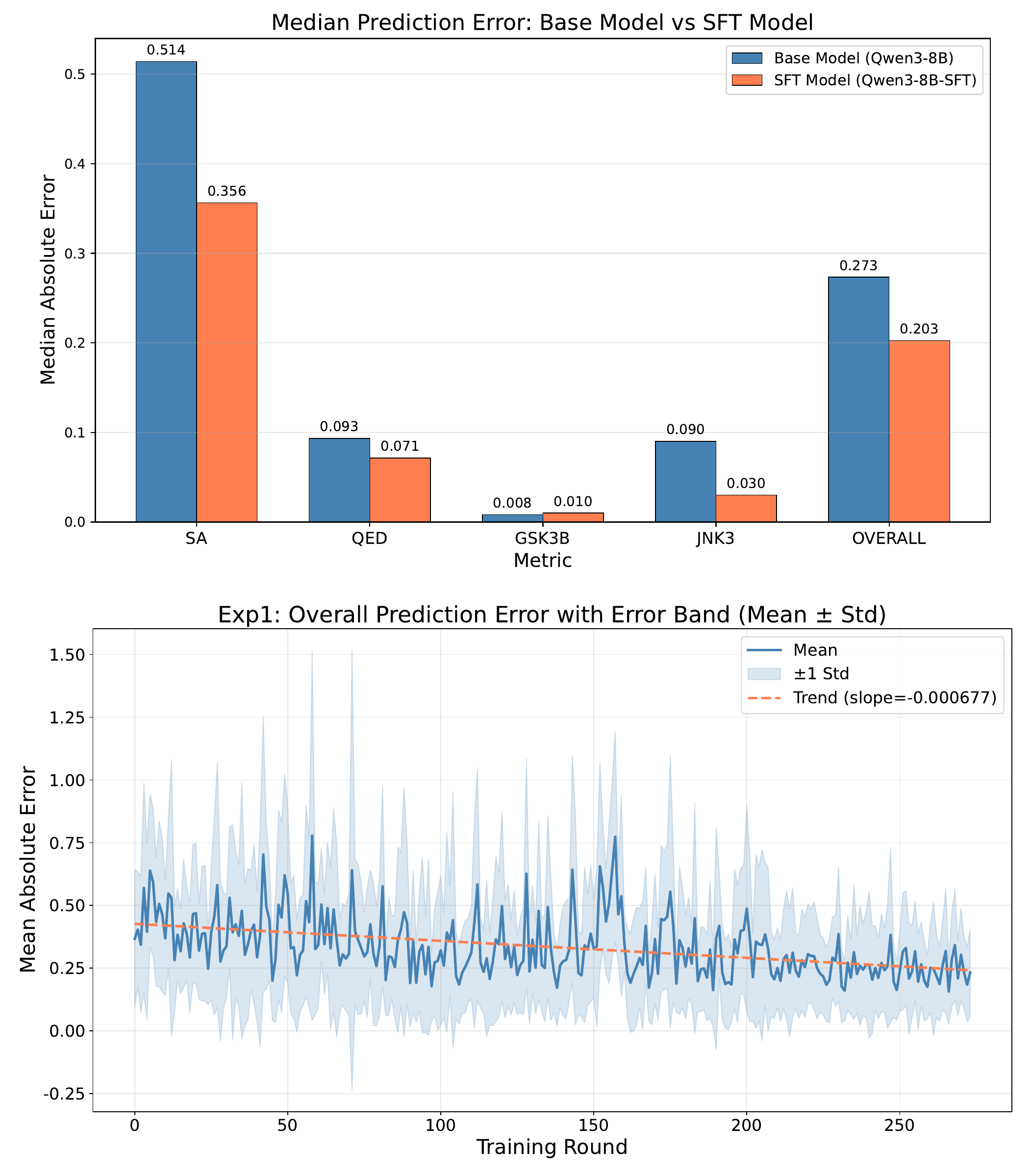}
  \caption{Prediction error decreases through warm start and interaction.}
  \label{f5}
  \vspace{-10pt}
\end{wrapfigure}

Table~\ref{ablation} isolates the effects of prediction, agentic refinement, warm start, reinforcement learning, and model scale. \textbf{Qwen3-8B} is a population-based LLM evolutionary baseline: at each generation, it receives selected parent candidates from the current population and proposes new offspring in a single round, similar to OpenEvolve-style LLM evolutionary search, but without tool use, explicit prediction, warm start, or reinforcement learning. \textbf{Qwen3-8B-Pred} adds pre-action prediction to this population-based one-shot generation setting, isolating the effect of prediction without agentic refinement. \textbf{Agent-Qwen3-8B} further adds tool-augmented multi-turn refinement, but removes explicit prediction, isolating the effect of agentic search. \textbf{Agent-Pred} combines tool use with predict-then-act reasoning, but uses neither SFT nor RL. \textbf{SFT-Agent-Pred} is the only variant with supervised warm start. \textbf{WMLLM-Evolve} is the full method with predict-then-act reasoning, agentic refinement, population-based evolution, and GRPO training, without SFT warm start. The remaining rows compare different backbones and model scales.
\begin{table}[t]
  \caption{Ablation study on multi-objective molecular optimization under \(3{,}000\) unique molecule evaluations.}
  \label{ablation}
  \centering
  \small
  \setlength{\tabcolsep}{2.5pt}
  \renewcommand{\arraystretch}{1.08}
  \newcolumntype{Y}{>{\centering\arraybackslash}X}
  \begin{tabularx}{\textwidth}{@{}lYYYYYY@{}}
    \toprule
    Model & \makecell{Avg.\\Top-1} & \makecell{Avg.\\Top-10} & \makecell{Top-10\\AUC} & Validity & Uniqueness & Diversity \\
    \midrule
    WMLLM-Evolve & \textbf{4.385$\pm$0.130} & \textbf{4.255$\pm$0.084} & \textbf{4.124$\pm$0.042} & 0.642$\pm$0.013 & 0.944$\pm$0.014 & 0.476$\pm$0.198 \\
    SFT-Agent-Pred & \underline{4.346$\pm$0.101} & \underline{4.251$\pm$0.033} & \underline{4.098$\pm$0.009} & 0.350$\pm$0.008 & \textbf{1.000$\pm$0.000} & 0.455$\pm$0.009 \\
    Agent-Pred & 4.329$\pm$0.092 & 4.248$\pm$0.053 & 4.095$\pm$0.032 & 0.389$\pm$0.014 & \textbf{1.000$\pm$0.000} & 0.333$\pm$0.192 \\
    Agent-Qwen3-8B & 4.071$\pm$0.106 & 4.025$\pm$0.094 & 3.871$\pm$0.049 & 0.304$\pm$0.044 & 0.994$\pm$0.003 & 0.479$\pm$0.071 \\
    Qwen3-8B & 4.088$\pm$0.201 & 4.026$\pm$0.138 & 3.827$\pm$0.178 & \underline{0.890$\pm$0.015} & 0.466$\pm$0.070 & 0.578$\pm$0.063 \\
    Qwen3-8B-Pred & 4.150$\pm$0.231 & 4.095$\pm$0.181 & 3.475$\pm$0.975 & \textbf{0.921$\pm$0.017} & 0.462$\pm$0.122 & 0.533$\pm$0.058 \\
    Qwen2.5-7B & 3.966$\pm$0.022 & 3.935$\pm$0.013 & 3.826$\pm$0.008 & 0.849$\pm$0.007 & 0.607$\pm$0.043 & 0.639$\pm$0.030 \\
    Qwen2.5-7B-Pred & 3.935$\pm$0.061 & 3.901$\pm$0.035 & 3.796$\pm$0.011 & 0.867$\pm$0.010 & 0.764$\pm$0.038 & \underline{0.657$\pm$0.031} \\
    Qwen3-32B & 4.172$\pm$0.171 & 4.111$\pm$0.147 & 3.869$\pm$0.188 & 0.656$\pm$0.048 & \underline{0.998$\pm$0.001} & 0.552$\pm$0.031 \\
    Qwen3-4B & 4.032$\pm$0.139 & 3.982$\pm$0.196 & 3.844$\pm$0.174 & 0.465$\pm$0.085 & \underline{0.998$\pm$0.000} & \textbf{0.663$\pm$0.051} \\
    \bottomrule
  \end{tabularx}
  \vskip -0.1in
\end{table}

The ablation results show three main trends. First, prediction improves one-shot generation on Qwen3-8B, increasing Avg. Top-1 from \(4.088\) to \(4.150\). Second, agentic refinement alone is insufficient: Agent-Qwen3-8B reaches \(4.071\), while Agent-Pred improves to \(4.329\), showing that tool use becomes more effective when the agent must predict before acting. Third, SFT-Agent-Pred slightly improves over Agent-Pred, indicating that warm start can help the predictive agent; however, WMLLM-Evolve achieves the best overall results without SFT, showing that reinforcement learning and population-based evolution can drive self-improvement from interaction.

The scaling rows further show that larger models alone do not explain the improvement. Qwen3-32B outperforms Qwen3-8B, but remains below Agent-Pred and WMLLM-Evolve. Thus, the main gain comes from predict-then-act optimization rather than model size alone.

\subsection{Prediction Fidelity and Self-Evolution}

A central hypothesis of WMLLM is that better prediction leads to better decisions. We therefore analyze prediction error before and during training.

Figure~\ref{f5} reports two analyses. The top panel shows that supervised warm start reduces median absolute prediction error across molecular properties and the overall score, explaining the strong performance of SFT-Agent-Pred in Table~\ref{ablation}. The bottom panel shows that prediction error decreases during interaction, indicating that WMLLM progressively refines its implicit estimate of evaluator feedback. Together with the ablation results, this supports the view that prediction fidelity is coupled with decision quality rather than being a passive auxiliary objective.

\subsection{Generalization to Additional Tasks}

We further evaluate Qwen3-8B on three additional black-box optimization tasks. \textsc{Circle Packing 26} maximizes the total radius of 26 non-overlapping circles in a unit square; \textsc{Sums Diffs} constructs an integer set optimizing sumset/difference-set statistics; and \textsc{Hadamard Det 29} searches for a \(29\times29\) \(\{\pm1\}\)-matrix with a large normalized determinant. All tasks use a 100-call evaluation budget, and WMLLM is trained with reinforcement learning without warm-start trajectories. Detailed task definitions, prompts, and evaluator protocols are provided in Appendix~\ref{app:generalization_tasks}.

\begin{wraptable}{r}{0.48\textwidth}
\vspace{-1.0em}
\centering
\caption{Generalization results on additional black-box optimization and agent environments.}
\label{tab:general_black_box}

\small
\resizebox{0.48\textwidth}{!}{
\begin{tabular}{lccc}
\toprule
Task & w/o Pred. & WMLLM & Gain \\
\midrule
Circle Packing 26 & 2.5027 & 2.6199 & +0.1172 \\
Sums Diffs & 1.0598 & 1.0934 & +0.0336 \\
Hadamard Det 29 & 0.8596 & 0.9311 & +0.0715 \\
\midrule
ALFWorld & $30.00{\pm}6.08$ & $43.59{\pm}4.02$ & +13.59 \\
\bottomrule
\end{tabular}
}
\vspace{-1.0em}
\end{wraptable}

As shown in Table~\ref{tab:general_black_box}, WMLLM improves the best score on all three black-box optimization tasks. These consistent gains suggest that predict-then-act world modeling is not limited to molecular representations, but can transfer to continuous geometric layouts, discrete additive-combinatorial constructions, and structured matrix search. We also evaluate ALFWorld as a preliminary out-of-domain test of general agent behavior.

\section{Conclusion}

We presented WMLLM, a self-evolving optimization-agent framework based on predict-then-act world modeling. By requiring the agent to predict candidate outcomes before acting, WMLLM turns prediction-observation discrepancy into a self-supervised signal for improving future decisions. Combined with agentic multi-turn interaction, population-based search, and trajectory-level optimization, this mechanism enables the policy and its implicit action-outcome model to improve through closed-loop feedback.

Experiments on multi-objective molecular optimization show that explicit prediction is a key driver of sample efficiency and final performance, outperforming gains from model scale or agentic refinement alone. Additional black box optimization tasks and preliminary ALFWorld results further suggest that predictive responsibility can improve agent behavior beyond the main molecular benchmark. Overall, WMLLM provides a practical step toward optimization agents that learn not only from rewards, but also from the errors in their own predictions.

\section{Limitations}

Our evidence mainly focuses on structured black box optimization, and broader validation on more diverse long-horizon agent environments remains future work. Although WMLLM improves sample efficiency, it still requires repeated evaluator feedback, so further reducing interaction cost is important for extremely expensive domains.



\newpage
\bibliographystyle{plainnat}
\bibliography{ref}

\newpage
\appendix
\setcounter{page}{1}
\vbox{
    \hrule height 4pt
    \vskip 0.25in
    \vskip -\parskip%
    \centering
    {\LARGE\bf WMLLM: Self-Evolving Optimization Agents via Predict-Then-Act World Modeling (Appendix)}
    \vskip 0.29in
    \vskip -\parskip
    \hrule height 1pt
}

\section{Additional Experimental Analysis}

\subsection{Evaluator Call Efficiency}

\begin{figure}[ht]
  \vskip 0.2in
  \begin{center}
    \centerline{\includegraphics[width=0.5\columnwidth]{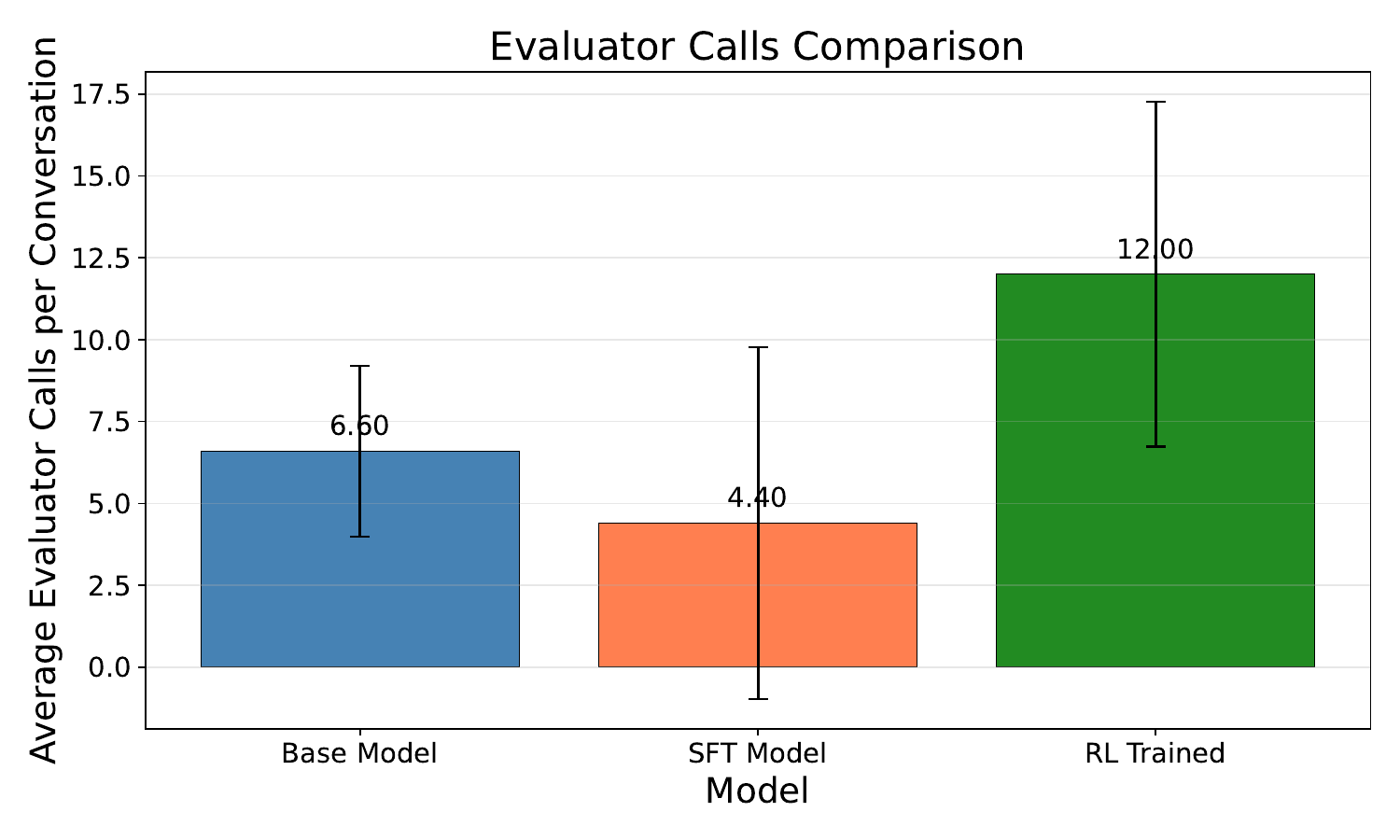}}
    \caption{
Evaluator call efficiency across training stages.
Bars report the average number of evaluator calls per conversation (mean $\pm$ std) for the base model, the SFT-initialized model, and the RL-trained agent. Warm-started models require fewer evaluator calls, reflecting more efficient decision-making enabled by improved predictive priors, while RL-trained agents exhibit longer and more variable interaction trajectories due to increased exploration.
}
    \label{f6}
  \end{center}
\end{figure}

This experiment examines how different training stages affect the agent’s use of expensive environment evaluations. Figure~\ref{f6} shows that the SFT-initialized model consistently requires fewer evaluator calls per conversation than the base model, indicating that causal warm-starting reduces redundant trial-and-error by providing more accurate predictive priors. In contrast, the RL-trained agent performs a larger and more variable number of evaluator calls, reflecting deeper refinement trajectories and more aggressive exploration. These results highlight a clear trade-off between interaction efficiency and exploration depth, and underscore the role of predictive initialization in improving sample-efficient decision making.

\section{Prompt Template}

\begin{promptlisting}{System Prompt for Agentic Molecular Design}
You are a multi-objective small-molecule design assistant.

Focus properties (direction):
- sa (minimize)
- drd2 (minimize)
- qed (maximize)
- gsk3b (minimize)
- jnk3 (maximize)

Final goal: overall score reported by evaluator >= 4.2.

Available tools:
1. analyzer - Decompose parents or candidates into fragments and functional groups.
2. evaluator - Returns property values and the aggregated score.

STRICT OUTPUT FORMAT - You MUST follow this exact format for EVERY candidate:

<candidate>SMILES_STRING</candidate>
<prediction>{"sa": NUMBER, "drd2": NUMBER, "qed": NUMBER,
             "gsk3b": NUMBER, "jnk3": NUMBER, "overall": NUMBER}</prediction>

CRITICAL RULES:
1. 5-SENTENCE LIMIT: Each response must contain NO MORE than 5 sentences
   of reasoning or explanation.
2. MANDATORY PREDICTION: You MUST output <prediction> with ALL 6 numeric
   values (sa, drd2, qed, gsk3b, jnk3, overall) for EVERY candidate BEFORE
   calling evaluator.
3. NUMERIC VALUES ONLY: All values in <prediction> must be numbers
   (e.g., 2.1, 0.001, 0.85).
4. NO REPEATED SMILES: Never evaluate the same molecule twice.
5. IMMEDIATE ACTION: After stating a candidate, immediately call evaluator.
   No planning without execution.

Example output format:

Rationale: [1-2 sentences max]. Self-check: all yes.
<candidate>COc1ccc(NC(=O)Cc2cccc3ccccc23)cc1</candidate>
<prediction>{"sa": 2.0, "drd2": 0.0002, "qed": 0.88,
             "gsk3b": 0.01, "jnk3": 0.04, "overall": 3.85}</prediction>
[tool call]

After evaluator returns, compare prediction vs actual in 1 sentence,
then immediately propose the next candidate with prediction.
\end{promptlisting}

\section{Cost Analysis}

We report the computational cost of WMLLM under the experimental setting used throughout the paper. All experiments are conducted with a fixed evaluation budget of 3{,}000 unique molecules. Training and optimization are performed on a cluster of 8 NVIDIA A100 GPUs with 80GB memory each.

Under this configuration, the average wall-clock training time is $3.53 \pm 0.24$ hours across multiple runs. The relatively low variance indicates stable training dynamics despite stochastic trajectory sampling and adaptive multi-turn interaction. Overall, these results suggest that WMLLM achieves its performance gains with moderate computational overhead, supporting its practicality for black-box optimization problems with expensive evaluations.

\section{Broader Impacts}

WMLLM aims to improve the sample efficiency of large language model agents for structured black-box optimization. Its potential positive impacts include reducing the number of expensive evaluator calls required in scientific discovery, molecular design, and automated engineering, thereby lowering computational and experimental costs. The predict-then-act mechanism may also improve the transparency of agent decisions by requiring the model to state expected outcomes before acting.
Potential negative impacts may arise if more efficient optimization agents are applied to harmful design spaces or sensitive decision-making domains without adequate constraints. In addition, inaccurate predictions may mislead the search process, so domain-specific safety filters, evaluator checks, and human oversight are important when deploying such systems.

\section{Details of Additional Black-Box Optimization Tasks}
\label{app:generalization_tasks}

This section provides the task definitions and evaluation protocols for the three additional black-box optimization tasks. These tasks are adapted from constructive optimization environments in which an agent proposes a candidate construction and receives feedback only through an external evaluator. For all three tasks, we use Qwen3-8B as the backbone model and impose the same budget of 100 evaluator calls. WMLLM and the \textit{w/o Pred.} baseline use the same task prompt, evaluator, backbone model, and evaluation budget; the only difference is whether the model is required to predict the evaluation outcome before proposing the candidate. We report the best score found within the evaluation budget.

\subsection{Circle Packing 26}

\textsc{Circle Packing 26} is a continuous geometric construction task. The goal is to place 26 circles inside the unit square \([0,1]^2\) while maximizing the sum of their radii. A candidate solution is represented as
\[
    \mathcal{C}=\{(x_i,y_i,r_i)\}_{i=1}^{26},
\]
where \((x_i,y_i)\) is the center of the \(i\)-th circle and \(r_i\geq 0\) is its radius. A valid solution must satisfy two types of constraints. First, every circle must lie entirely inside the unit square:
\[
    r_i \leq x_i \leq 1-r_i,\qquad
    r_i \leq y_i \leq 1-r_i.
\]
Second, any two circles must be non-overlapping:
\[
    \sqrt{(x_i-x_j)^2+(y_i-y_j)^2} \geq r_i+r_j,
    \qquad \forall i\neq j.
\]
The evaluator checks these constraints and computes the objective
\[
    S(\mathcal{C})=\sum_{i=1}^{26} r_i.
\]
Invalid candidates are rejected by the evaluator, while valid candidates are scored by their true radius sum. For WMLLM, the model is asked to predict task-specific quantities such as the expected radius sum, validity, and possible failure causes before submitting the candidate. This task tests whether the model can acquire useful geometric priors about boundary constraints, pairwise collisions, and local layout improvements.

\subsection{Sums Diffs}

\textsc{Sums Diffs} is a discrete additive-combinatorial construction task. The agent must construct a finite integer set \(A\). Given \(A\), its sumset and difference set are defined as
\[
    A+A=\{a_i+a_j: a_i,a_j\in A\},
    \qquad
    A-A=\{a_i-a_j: a_i,a_j\in A\}.
\]
The task objective is
\[
    C(A)=
    \frac{\log(|A+A|/|A|)}
    {\log(|A-A|/|A|)}.
\]
The evaluator verifies that the submitted object is a valid integer set under the required size and value constraints, then recomputes \(|A+A|\), \(|A-A|\), and \(C(A)\) from scratch. Therefore, the model cannot obtain a high score by reporting an incorrect objective value. The final score is the recomputed value of \(C(A)\), and larger values are better.

For WMLLM, the model predicts quantities such as \(|A+A|\), \(|A-A|\), the resulting \(C(A)\), and candidate validity before receiving evaluator feedback. This task is useful for testing prediction-guided decision making because the candidate representation is simple, the evaluator feedback is deterministic, and the quality of a candidate depends on nontrivial global set statistics rather than local syntactic validity alone.

\subsection{Hadamard Det 29}

\textsc{Hadamard Det 29} is a structured matrix construction task. The goal is to construct a \(29\times29\) matrix
\[
    H\in\{-1,+1\}^{29\times29}
\]
that maximizes the absolute determinant \(|\det(H)|\). Since determinant magnitude grows rapidly with matrix size, the evaluator reports a normalized determinant ratio:
\[
    R(H)=\frac{|\det(H)|}{29^{29/2}},
\]
where \(29^{29/2}\) is the Hadamard upper bound for matrices with entries in \(\{-1,+1\}\). The evaluator first checks that the submitted matrix has the correct shape and contains only \(\pm1\) entries. It then computes the determinant and returns the normalized ratio \(R(H)\) as the task score.

For WMLLM, the model predicts the expected determinant ratio and validity of the proposed matrix before evaluation. Compared with the previous two tasks, this environment is more structurally demanding: the effect of changing a single matrix entry is global and depends on row correlations, near-orthogonality, and the conditioning of the full matrix. As a result, this task evaluates whether predict-then-act modeling can transfer to high-dimensional discrete matrix search rather than only to geometric or set-based constructions.

\subsection{Evaluation Protocol}

Across all three tasks, the evaluator is treated as a black box: the model proposes a candidate, the evaluator verifies validity and recomputes the true score, and the search process continues using the returned feedback. The \textit{w/o Pred.} baseline follows the same optimization protocol but removes the explicit prediction step and the prediction-error learning signal. WMLLM, in contrast, must first produce a task-specific prediction of the candidate's evaluation outcome and then propose the candidate. This design allows us to test whether improving the model's outcome prediction also improves its subsequent optimization decisions under a fixed evaluation budget.

\section{Implementation Details and Hyperparameters}
\label{app:implementation}

This section provides additional implementation details for WMLLM, including population construction, warm-start data, GRPO training, evaluator budget, and random seeds. Unless otherwise stated, these settings are used for the main multi-objective molecular optimization experiments.

\begin{table}[h]
\centering
\caption{Implementation details and hyperparameters used in WMLLM.}
\label{tab:implementation_details}
\small
\setlength{\tabcolsep}{5pt}
\renewcommand{\arraystretch}{1.08}
\begin{tabular}{ll}
\toprule
Hyperparameter / Setting & Value \\
\midrule
Population size \(N\) & 50 \\
Initial candidate pool & 100 random molecules \\
Parent number per prompt & 3 \\
Parent prompts per generation & 10 \\
Offspring number & Variable, determined by valid model outputs \\
Top-score selection ratio & 0.5 \\
Pareto selection ratio & 0.5 \\
Hybrid population construction & 50\% top-score selection + 50\% Pareto selection \\
Warm-start data size & 700 total examples \\
Warm-start train / validation split & 630 / 70 \\
GRPO group size & 8 sampled trajectories per prompt \\
Prediction reward weight \(\alpha\) & 0.2 \\
GRPO actor learning rate & \(1\times 10^{-6}\) \\
Rollout batch size & 10 \\
PPO mini-batch size & 4 \\
Micro-batch size per GPU & 1 \\
Number of RL iterations & 300 \\
Maximum prompt length & 2560 tokens \\
Maximum response length & 10240 tokens \\
Maximum assistant turns & 20 \\
Evaluator call budget & 3000 unique molecule evaluations \\
Random seeds & 42, 43, 44, 45, 46 \\
\bottomrule
\end{tabular}
\end{table}

\paragraph{Population construction.}
WMLLM maintains a population of \(N=50\) evaluated candidates. At each generation, the model receives a subset of three parent molecules as context and proposes new offspring through either single-round generation or agentic multi-turn refinement, depending on the variant. The number of offspring is not fixed in advance, because each prompt may produce a variable number of valid candidates during interaction. All candidates submitted to the evaluator consume budget.
WMLLM is evaluated as an online optimization process: population update, evaluator feedback, and GRPO training occur within the same budgeted search run. We do not carry over an additional hidden test-time population beyond the reported evaluation budget.

\paragraph{Hybrid population update.}
To balance exploitation and diversity, we use a hybrid update rule. Half of the next population is selected by the scalar overall score, and the other half is selected by Pareto ranking over the objective vector. Concretely, the top-score selection ratio is \(0.5\), and the Pareto selection ratio is \(0.5\). This design keeps high-performing candidates while preserving diverse trade-offs among objectives.

\paragraph{Warm start.}
The supervised warm-start dataset contains 700 score-prediction examples, split into 630 training examples and 70 validation examples. Warm start is used only in the \textsc{SFT-Agent-Pred} variant. The full \textsc{WMLLM-Evolve} method does not use SFT warm start, so its improvement comes from interaction, population evolution, and GRPO training.

\paragraph{GRPO training.}
For each prompt, GRPO samples a group of 8 trajectories and computes group-normalized advantages within the same prompt group. The reward combines evaluator reward and prediction reward with prediction weight \(\alpha=0.2\), i.e., the total reward is \(0.8\) times the evaluator reward plus \(0.2\) times the prediction reward. The actor learning rate is \(1\times 10^{-6}\). We use a rollout batch size of 10, PPO mini-batch size of 4, and micro-batch size 1 per GPU. The number of RL iterations is 300.

\paragraph{Trajectory and budget limits.}
The maximum prompt length is 2560 tokens and the maximum response length is 10240 tokens. The maximum number of assistant turns is 20. For the main molecular optimization benchmark, all methods are evaluated under a budget of 3000 unique molecule evaluations. Variable-depth agentic refinement is allowed, but every evaluator call made during refinement is counted toward the same budget.

\paragraph{Random seeds.}
All main results are averaged over five random seeds, \(42\) to \(46\). We report mean and standard deviation across these seeds.

\end{document}